\documentclass[10pt,twocolumn,letterpaper]{article}

\usepackage[table]{xcolor}
\definecolor{lightgreen}{RGB}{240, 255, 240}  
\definecolor{headergray}{gray}{0.96}
\usepackage{pifont}  
\usepackage[pagenumbers]{cvpr} 

\definecolor{cvprblue}{rgb}{0.21,0.49,0.74}
\usepackage[pagebackref,breaklinks,colorlinks,allcolors=cvprblue]{hyperref}

\usepackage{multirow}
\usepackage[table]{xcolor}
\definecolor{lightgray}{gray}{0.94}
\title{Screening Is Effective for Visual Recognition}

\author{
Shunya Shimomura\\
Meijo University\\
\and
Kazuhiro Hotta\\
Meijo University\\
}

\begin{document}
\maketitle
\begin{abstract}
Vision Transformer (ViT) has been widely used as a powerful framework for modeling global dependencies among image patches. However, its core component, self-attention assigns softmax-normalized relative weights to all patches, making it difficult to evaluate the relevance between patches independently. In visual recognition, images often contain many background or redundant patches, yet self-attention cannot explicitly reject such irrelevant patches, which may introduce unnecessary information into feature aggregation. To address this limitation, Screening has been proposed in the field of language modeling, where the relevance of each token is independently evaluated based on query–key similarity and low-relevance tokens are explicitly excluded through thresholding.
In this work, we propose \textbf{VisionScreen}, a new vision model that extends Screening mechanism to visual recognition. VisionScreen treats image patches as tokens arranged on a two-dimensional grid and extends absolute relevance estimation based on query–key similarity to the two-dimensional spatial domain. This allows each patch to selectively aggregate only content-wise and spatially relevant patches without relying on competition among patches.
Experiments on image classification benchmarks demonstrate that the proposed method outperforms conventional ViT. These results suggest that Screening can be effective for visual recognition, offering an alternative to relative feature aggregation based on softmax attention.

\end{abstract}

\section{Introduction}
\label{sec:intro}
Vision Transformer (ViT) has become one of the representative architectures in visual recognition~\cite{dosovitskiy2020vit, kirillov2023segment, radford2021clip}. ViT divides an image into a sequence of patches and processes each patch as a token, enabling it to capture long-range dependencies within an image in a manner different from convolutional neural networks. In particular, self-attention can directly model relationships between distant image regions, and has shown strong performance across various visual recognition tasks, including image classification, object detection, and semantic segmentation~\cite{xie2021segformer, liu2021swin, touvron2021deit, he2022masked, kirillov2023segment}. Thus, patch-wise interactions based on self-attention have become an important component of modern vision models.

However, self-attention in ViT aggregates features based on relative weights normalized by softmax. Specifically, the weight assigned to each key patch for a given query patch is not determined independently by the query–key pair itself, but is instead determined through competition with the other key patches considered at the same time. This property makes it difficult to interpret the relevance between patches as an absolute measure. Moreover, because softmax normalization assigns non-negative weights to all keys, it cannot explicitly reject patches with low relevance. In visual recognition, an input image often contains not only the target object but also many patches corresponding to background regions, textures, and redundant local patterns that are not necessarily required for recognition. Therefore, self-attention, which relatively weights and aggregates all patches, may allow redundant or irrelevant information to be mixed into the feature representation, potentially degrading recognition performance and interpretability.

To address such limitations of softmax attention, a mechanism called Screening has been proposed in the field of language modeling~\cite{nakanishi2026screening}. Screening first normalizes queries and keys, computes bounded query–key similarities, and then applies thresholding to these similarities to independently evaluate the relevance of each key. While softmax attention redistributes probability mass across all keys, Screening defines an absolute relevance value for each query--key pair and assigns zero relevance to pairs below the threshold, thereby removing their contribution during value aggregation.
This property enables Screening to explicitly reject irrelevant tokens and selectively aggregate only relevant tokens. Another important difference from self-attention is that Screening does not need to forcefully assign weight to any key when no relevant key exists, allowing it to represent the absence of relevant information.

In this work, we propose VisionScreen, a new vision model that adapts the Screening mechanism to two-dimensional image patch grids for visual recognition. VisionScreen treats image patches as tokens arranged on a two-dimensional grid and computes absolute relevance based on query–key similarity between patches. Furthermore, by extending the distance-based selection mechanism in Screening to two-dimensional space, each patch can selectively aggregate patches that are both content-wise relevant and spatially appropriate. As a result, the proposed method aims to learn feature representations that are less affected by redundant or irrelevant patches by judging the relevance of each patch independently, without relying on competition among other patches. In contrast to the original Screening mechanism, which selectively aggregates past tokens along a causal one-dimensional sequence, visual recognition is a non-causal setting in which all patches are observed simultaneously and each patch has two-dimensional neighborhood relationships. Therefore, we introduce Screening that considers relative positions between image patches, extending selection along the sequence direction to selection in the spatial direction, thereby realizing feature aggregation suited to the spatial structure of image data.

In our experiments, we evaluate the proposed method on image classification benchmarks and show that it achieves effective recognition performance compared with conventional ViT. These results suggest that selective patch aggregation based on Screening can be effective for visual recognition as an alternative to relative information aggregation based on softmax attention.

\section{Related Work}
\label{sec:related}

\subsection{Vision Transformers and Softmax Attention}
ViT models images as sequences of patches and captures global dependencies via self-attention~\cite{dosovitskiy2020vit}. Subsequent variants such as DeiT~\cite{touvron2021deit}, Swin Transformer~\cite{liu2021swin}, and MAE~\cite{he2022masked} have further improved performance, demonstrating the effectiveness of attention-based architectures in vision.

However, these models rely on softmax attention, where patch interactions are computed as relative weights normalized over all keys. As a result, each query--key relevance is not evaluated independently, and low-relevance patches cannot be explicitly rejected due to the non-zero weighting of all keys~\cite{nakanishi2026screening}. In contrast, our method introduces Screening to enable independent relevance evaluation and explicit exclusion of low-relevance patches.

\subsection{Vision Models Replacing Attention}

Although self-attention is a powerful token mixing mechanism, several vision models have explored alternatives to attention for aggregating information across patches. MLP-Mixer performs token mixing and channel mixing using only multilayer perceptrons, without convolution or attention, and achieves competitive performance in image recognition~\cite{tolstikhin2021mlpmixer, touvron2022resmlp}. PoolFormer further argues that the general MetaFormer structure, rather than attention itself, is a key factor for strong visual representation learning, and shows that even a simple pooling operation can serve as an effective token mixer~\cite{yu2022metaformer}. These studies suggest that self-attention is not necessarily an indispensable component for visual recognition, and that alternative token mixing mechanisms can also learn effective visual features.

However, attention-replacement methods such as MLP-Mixer and PoolFormer mainly focus on simplifying or improving token mixing, rather than independently evaluating the relevance of each patch based on query--key relationships. In contrast, our method computes absolute relevance based on query--key similarity using Screening and explicitly removes low-relevance patches through thresholding. Therefore, our work differs from existing attention-replacement approaches in that it does not merely replace attention with a simpler operation, but addresses the relative-weighting limitation of softmax attention and introduces a selective patch aggregation mechanism based on absolute relevance.

\section{Method}

\begin{figure}
\centering
\includegraphics[width=1\linewidth]{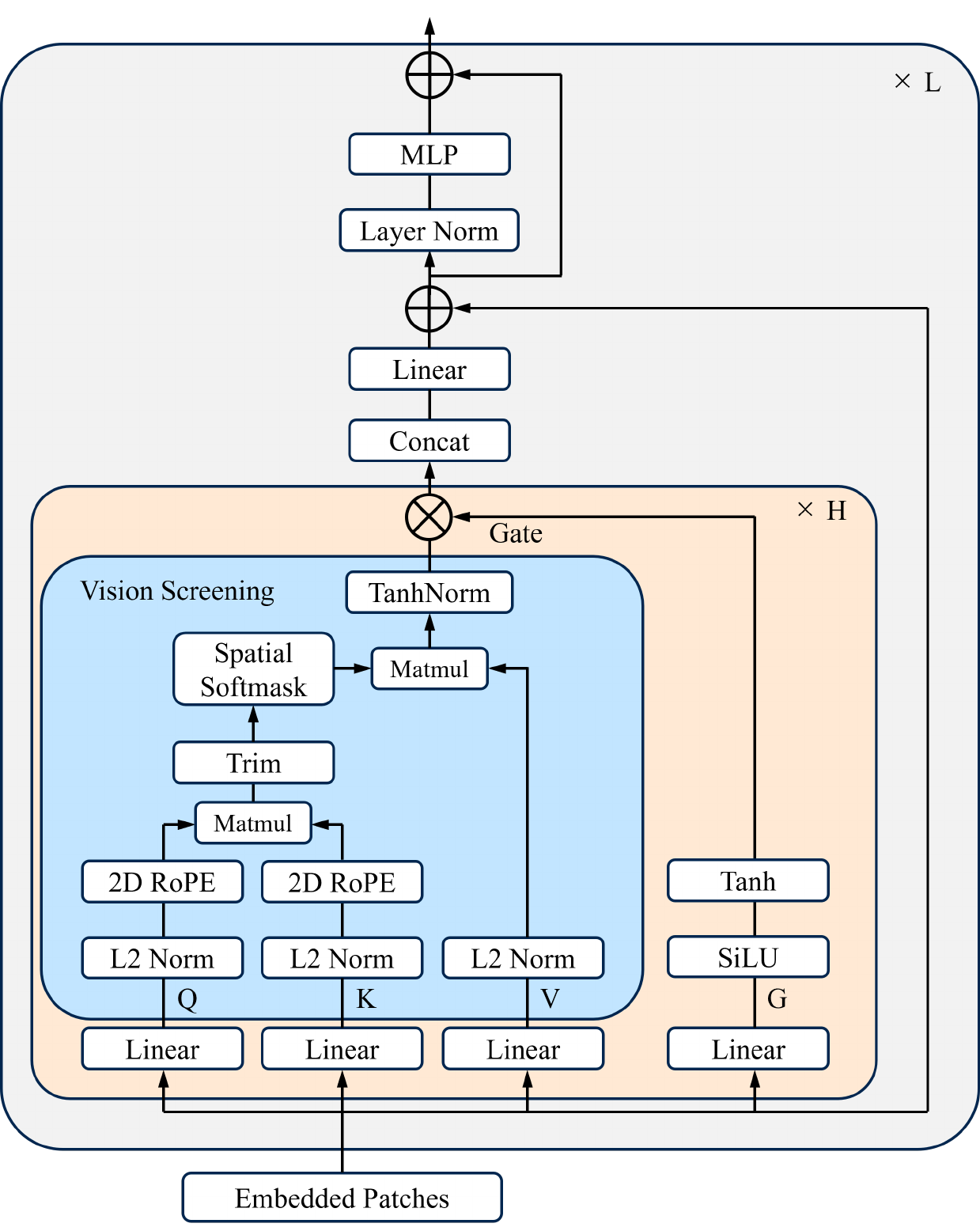}
\caption{
Overview of VisionScreen.
}
\label{method}
\end{figure}
\begin{table*}
\centering
\caption{
Quantitative comparison between ViT-Tiny/16 and VisionScreen.
We report top-1 accuracy on ImageNet-1k and CIFAR-100, together with the number of model parameters. The best values are in bold.
}
\label{tab:main_results}
\begin{tabular}{lccc}
\toprule
Model & Param. (M) & ImageNet-1k Top-1 (\%) & CIFAR-100 Top-1 (\%) \\
\midrule 
ViT-Tiny/16 & 5.7 & 68.1 & 50.3 \\
VisionScreen & \textbf{5.4} & \textbf{72.5} & \textbf{52.4} \\
\bottomrule
\end{tabular}
\end{table*}

\subsection{Overview}

We adapt the Screening mechanism to visual recognition.
Our goal is not to redesign the Screening operation itself, but to make it suitable for non-causal two-dimensional image patch grids.
As shown in Fig.~\ref{method}, we replace the self-attention module in a ViT block with a two-dimensional Screening module.

Given an input image, we first divide it into non-overlapping patches and embed them into patch tokens following the standard ViT pipeline.
Let $\mathbf{X} = [\mathbf{x}_1, \ldots, \mathbf{x}_N] \in \mathbb{R}^{N \times d}$ denote the patch tokens, where each patch token has a two-dimensional grid coordinate $\mathbf{p}_n = (p_n^x, p_n^y)$.
The proposed module follows the original Screening mechanism for query, key, value, and gate projections, unit-length normalization, Trim transform, TanhNorm, and gated aggregation.
The main modifications are the introduction of axial two-dimensional RoPE and the extension of the distance-aware softmask to two-dimensional space.

Unlike the causal processing of the original Screening mechanism, VisionScreen allows each patch to compute relevance over all patches in the image grid.
This non-causal setting is suitable for image recognition, where all patches are available simultaneously.

\subsection{Extensions for 2D images}
\subsubsection{2D RoPE}

The original Screening mechanism is designed for one-dimensional token sequences.
In contrast, image patches are arranged on a two-dimensional grid, and their spatial relationships are important for visual recognition.
To incorporate such spatial structure into query--key similarity, we adopt axial two-dimensional Rotary Position Embedding (2D RoPE), following RoPE for ViT~\cite{heo2024rope}, which extends RoPE~\cite{su2024roformer} to vision models.

Let $\mathbf{p}_n=(p_n^x,p_n^y)$ denote the two-dimensional coordinate of the $n$-th patch, where $p_n^x$ and $p_n^y$ correspond to the horizontal and vertical positions, respectively.
For each head, we first obtain normalized query and key vectors $\bar{\mathbf{q}}_n$ and $\bar{\mathbf{k}}_n$ as in the original Screening mechanism.
We divide their feature dimensions into two equal groups corresponding to the horizontal and vertical axes.
Within each group, every pair of feature dimensions is rotated according to the patch coordinate along the corresponding axis.

For an axis $a \in \{x,y\}$, the rotation applied to the $t$-th pair of feature dimensions is defined as
\begin{equation}
\mathbf{R}(p_n^a,t)
=
\begin{bmatrix}
\cos(\theta_t p_n^a) & -\sin(\theta_t p_n^a) \\
\sin(\theta_t p_n^a) & \cos(\theta_t p_n^a)
\end{bmatrix},
\label{eq:axial_rope}
\end{equation}
where $p_n^a$ denotes the coordinate of the $n$-th patch along axis $a$.
The frequency is defined as
\begin{equation}
\theta_t
=
10000^{-t/(d_h/4)},
\quad
t \in \{0,1,\ldots,d_h/4-1\},
\label{eq:rope_frequency}
\end{equation}
where $d_h$ denotes the head dimension of the query and key vectors.

Let $\bar{\mathbf{q}}_{n,a,t}$ and $\bar{\mathbf{k}}_{n,a,t}$ denote the $t$-th pair of query and key dimensions assigned to axis $a$.
The position-encoded components are obtained as
\begin{equation}
\tilde{\mathbf{q}}_{n,a,t}
=
\bar{\mathbf{q}}_{n,a,t}
\mathbf{R}(p_n^a,t)^\top,
\quad
\tilde{\mathbf{k}}_{n,a,t}
=
\bar{\mathbf{k}}_{n,a,t}
\mathbf{R}(p_n^a,t)^\top.
\end{equation}
The rotated components for the two axes together form the position-encoded query and key vectors $\tilde{\mathbf{q}}_n$ and $\tilde{\mathbf{k}}_n$.
The similarity between query patch $n$ and key patch $m$ is then computed as
\begin{equation}
s_{nm}
=
\tilde{\mathbf{q}}_n
\tilde{\mathbf{k}}_m^\top.
\end{equation}
Thus, the query--key similarity incorporates relative spatial information along both horizontal and vertical axes.

\subsubsection{Spatial Softmask}

The original Screening mechanism uses a distance-aware softmask defined along a one-dimensional causal sequence.
We extend this softmask to two-dimensional image space by using the Euclidean distance between patch coordinates.
For two patches at $\mathbf{p}_i$ and $\mathbf{p}_j$, we define
\begin{equation}
d_{ij} = |\mathbf{p}_i - \mathbf{p}_j|_2 .
\end{equation}
Each head has a learnable screening window
\begin{equation}
w = \exp(s_w) + 1,
\label{eq:screening_window}
\end{equation}
where $s_w$ is a learnable scalar.
Following the original Screening mechanism~\cite{nakanishi2026screening}, we use a cosine function to smoothly decrease the spatial weight as the distance from the query patch increases.
The weight is equal to one when the two patches are at the same position and gradually decreases to zero at the boundary of the screening window.
The two-dimensional spatial softmask is then defined as
\begin{equation}
m_{ij} =
\begin{cases}
\frac{1}{2}
\left(
\cos \left( \frac{\pi d_{ij}}{w} \right) + 1
\right), 
& d_{ij} < w, \\
0,
& \mathrm{otherwise}.
\end{cases}
\label{eq:2d_softmask}
\end{equation}

This formulation allows each head to learn a spatial screening range over the image plane.
To compute the final relevance score, we first apply the same Trim transform as in the original Screening mechanism to the query--key similarity $s_{ij}$
\begin{equation}
\alpha_{ij} = \operatorname{Trim}(s_{ij}).
\end{equation}
We then apply the spatial softmask to the resulting content-based relevance score
\begin{equation}
\alpha_{ij}^s = \alpha_{ij} m_{ij}.
\end{equation}
Unlike softmax attention, $\alpha_{ij}^s$ is not normalized over keys.
Therefore, each patch is selected based on its absolute relevance to the query patch rather than through competition with other patches.



\begin{figure}
\centering
\begin{subfigure}{1\linewidth}
\centering
\includegraphics[width=\linewidth]{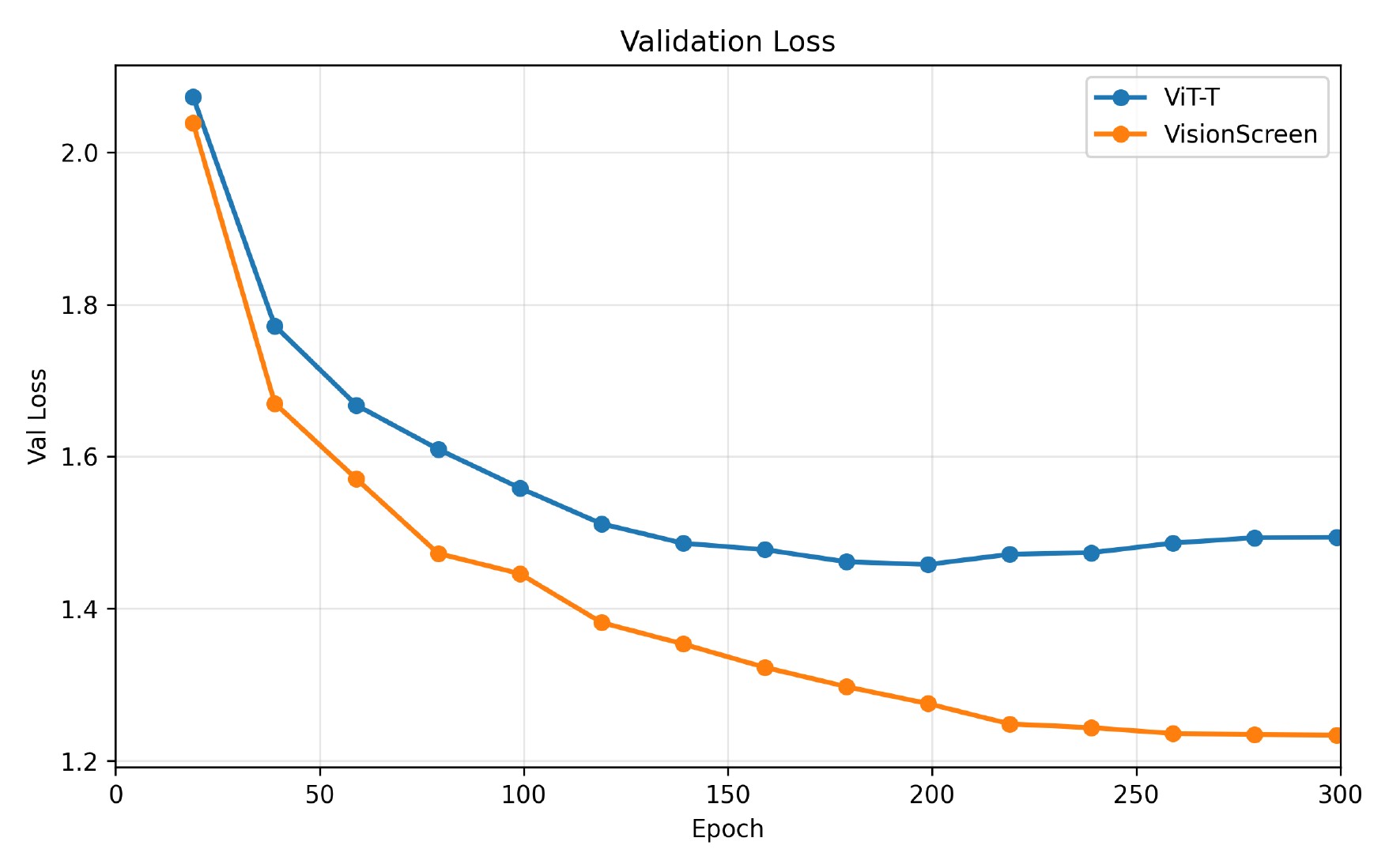}
\caption{ImageNet-1k}
\label{val_loss_in1k}
\end{subfigure}

\begin{subfigure}{1\linewidth}
\centering
\includegraphics[width=\linewidth]{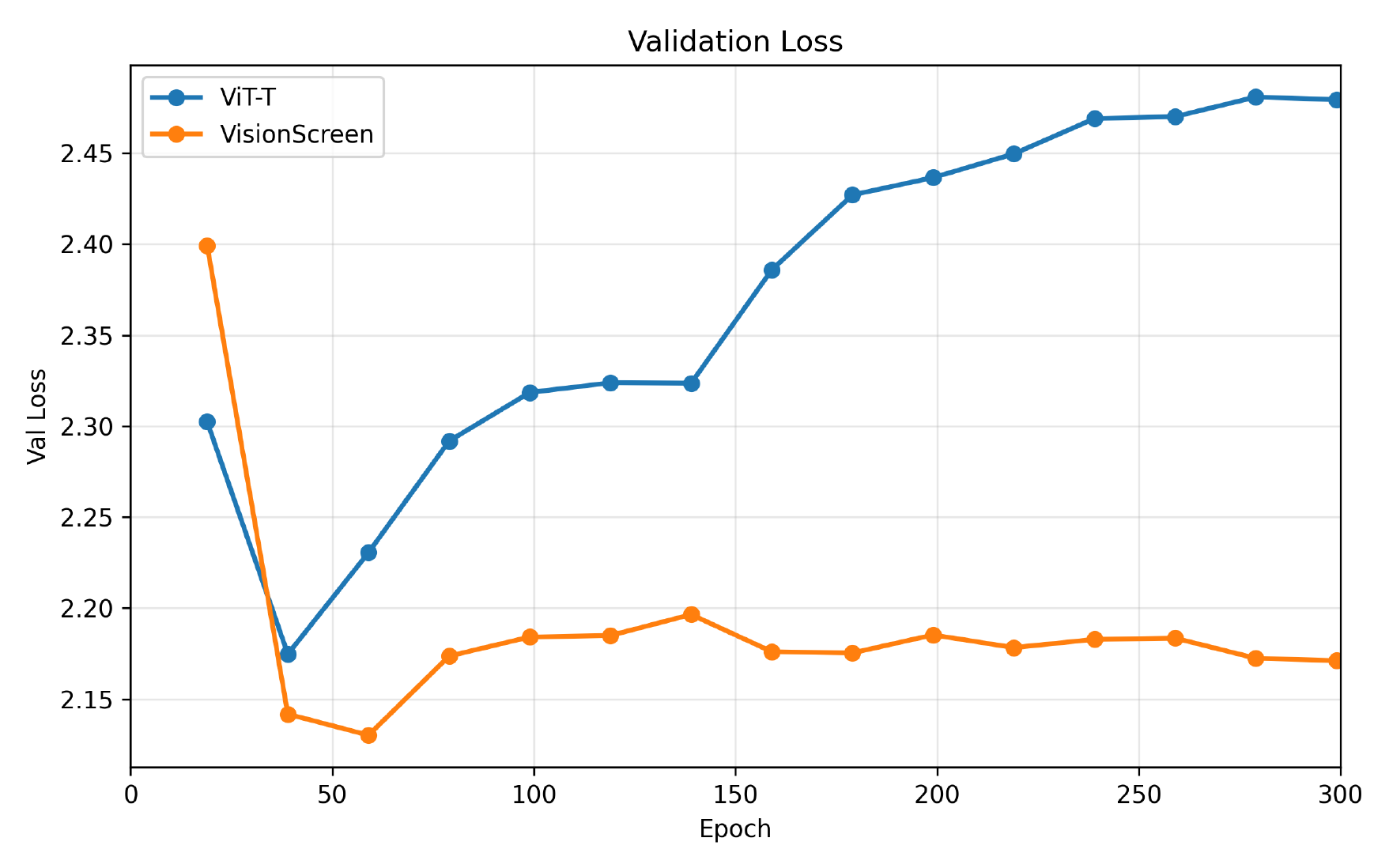}
\caption{CIFAR-100}
\label{val_loss_cifar100}
\end{subfigure}

\caption{
Validation loss curves of ViT-Tiny/16 and VisionScreen on ImageNet-1k and CIFAR-100.
}
\label{fig:val_loss}
\end{figure}

\section{Experiments}
\label{sec:experiments}

\subsection{Experimental Setup}

We evaluate VisionScreen on image classification benchmarks. As a baseline, we use ViT-Tiny/16 with an input resolution of $224 \times 224$. Both ViT-Tiny/16 and VisionScreen follow the same overall model scale: 12 layers, 3 heads, an embedding dimension of 192, and a patch size of $16 \times 16$.
We conduct experiments on ImageNet-1k~\cite{deng2009imagenet} and CIFAR-100~\cite{krizhevsky2009cifar}.
All models are trained from scratch on each dataset.
For ImageNet-1k, we train models with a global batch size of 1024, while for CIFAR-100 we use a global batch size of 512.
All models are trained for 300 epochs, with the first 5 epochs used for learning-rate warmup followed by cosine learning-rate scheduling. We use AdamW as the optimizer.
To isolate the effect of the architecture, we avoid advanced data augmentation and regularization techniques.
Following simple training settings used in prior work~\cite{kolesnikov2020BiT, steiner2021train, dosovitskiy2020vit}, we only apply random resized cropping and random horizontal flipping during training.

\subsection{Quantitative Comparison with ViT}

We compare VisionScreen with the ViT-Tiny/16 baseline on ImageNet-1k and CIFAR-100.
Table~\ref{tab:main_results} shows the top-1 classification accuracy and the number of model parameters.
VisionScreen achieves higher accuracy than ViT on both datasets while using fewer parameters.
On ImageNet-1k, VisionScreen improves top-1 accuracy from 68.1\% to 72.5\%, corresponding to a gain of 4.4 percentage points.
On CIFAR-100, VisionScreen improves top-1 accuracy from 50.3\% to 52.4\%, showing a gain of 2.1 percentage points.
These results indicate that Screening-based selective patch aggregation is effective for visual recognition and can provide better recognition performance than softmax attention under a comparable model scale.

In addition to classification accuracy, we also compare the validation loss curves during training.
Figure~\ref{fig:val_loss} shows the validation loss on ImageNet-1k and CIFAR-100.
On both datasets, VisionScreen achieves a lower validation loss than the ViT baseline throughout the later stages of training.
This trend is consistent with the top-1 accuracy results in Table~\ref{tab:main_results}, suggesting that the improvement of VisionScreen is not limited to the final classification accuracy but is also reflected in the optimization behavior.
These results indicate that the proposed Screening-based patch aggregation can learn more effective visual representations than softmax attention under the same training setting.

VisionScreen also uses fewer parameters than the ViT-Tiny/16 baseline.
This reduction mainly comes from two design choices.
First, VisionScreen uses axial 2D RoPE instead of learnable absolute position embeddings.
Second, while ViT-Tiny/16 uses a query and key dimension of 64 for each head, VisionScreen uses a dimension of 16, following the original Screening configuration~\cite{nakanishi2026screening}.
Since Screening computes relevance from unit-normalized query and key vectors, the relevance score is based on bounded cosine similarity rather than a high-dimensional dot product followed by softmax normalization.
Therefore, we consider that VisionScreen can effectively estimate patch relevance even with lower-dimensional query and key representations.

\subsection{Visualization of Patch Interactions}

To further analyze the behavior of VisionScreen, we visualize token-to-token interaction maps and compare them with those of ViT.
For ViT, we visualize the softmax attention weights averaged over all heads in the final layer.
For VisionScreen, we visualize the relevance scores computed from query--key similarity before value aggregation, also averaged over all heads in the final layer.

\begin{figure}
\centering
\includegraphics[width=0.95\linewidth]{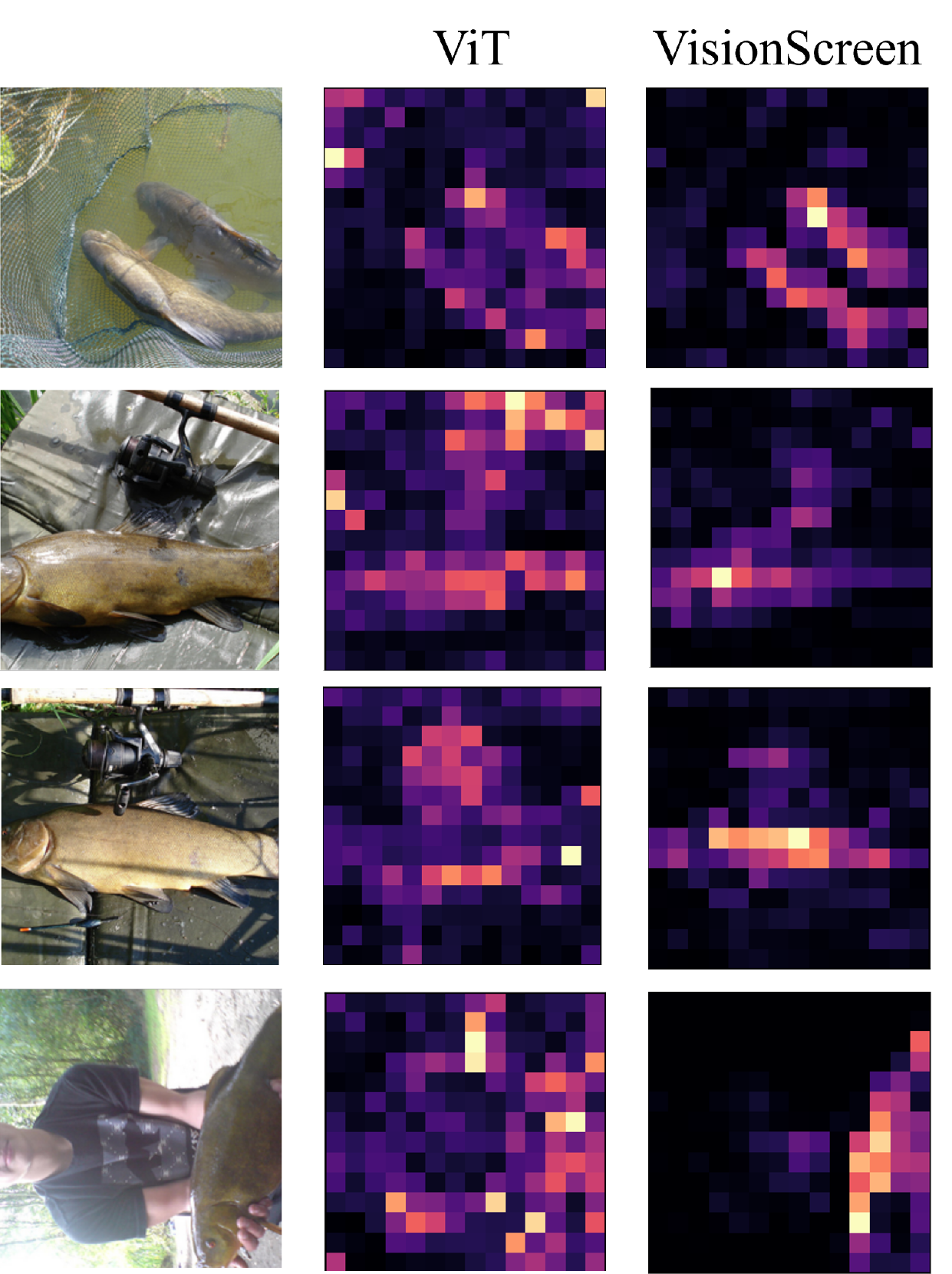}
\caption{
Visualization of CLS-to-patch interactions.
For ViT, we visualize softmax attention weights.
For VisionScreen, we visualize relevance scores before value aggregation.
}
\label{fig:attention_cls}
\end{figure}

\begin{figure}
\centering
\includegraphics[width=0.95\linewidth]{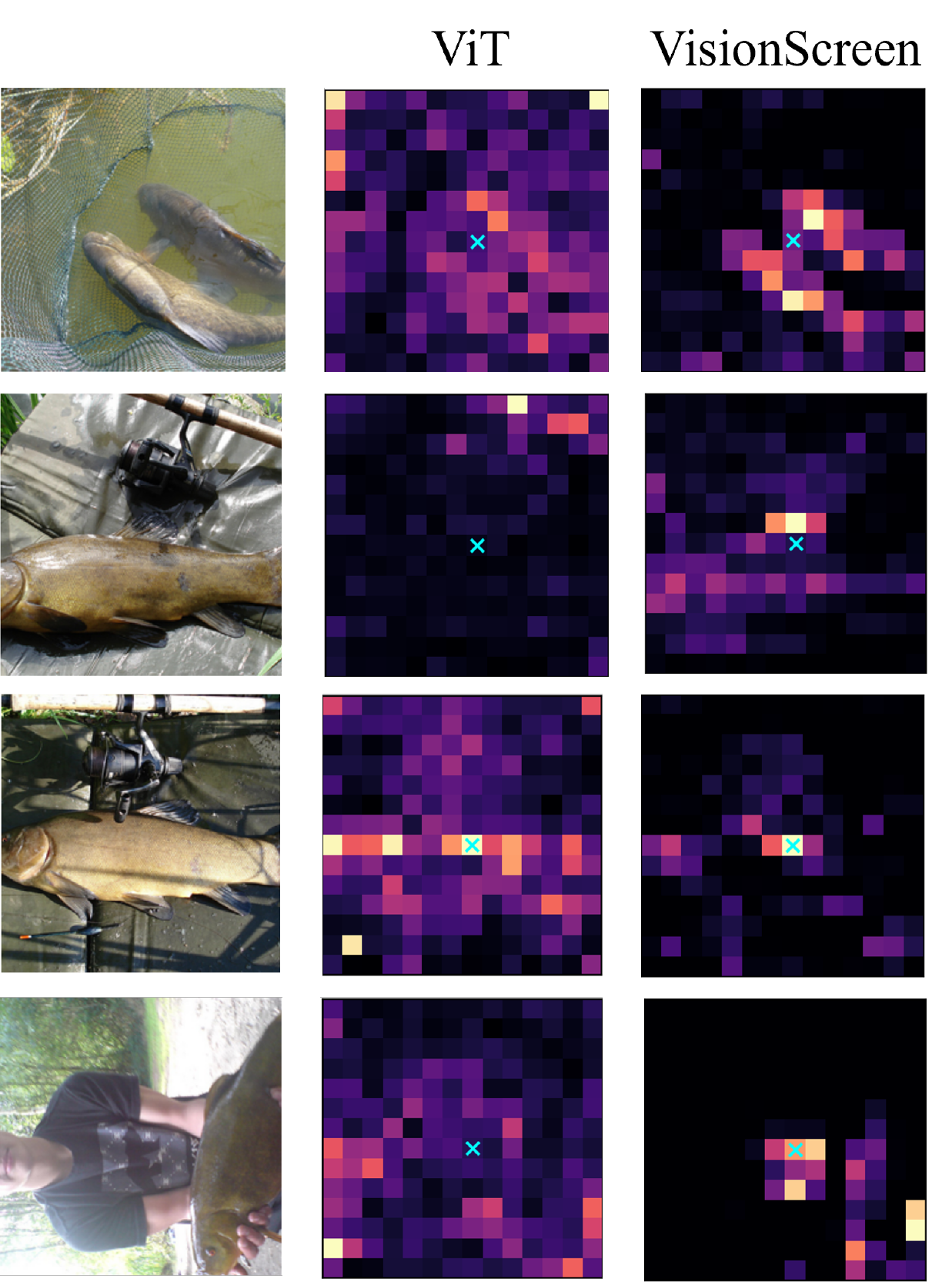}
\caption{
Visualization of interactions from a specific patch (marked with ×) to all patches.
}
\label{fig:attention_center}
\end{figure}

We consider two types of visualizations.
First, Fig.~\ref{fig:attention_cls} visualizes the interaction from the classification token (CLS token) to all patch tokens.
This visualization shows which image regions contribute to the global representation used for classification.
Second, Fig.~\ref{fig:attention_center} visualizes the interaction from a center patch to all patch tokens.
The selected center patch is marked with a light-blue cross in the figure.
This visualization allows us to observe which regions the selected patch refers to when aggregating information from other patches.

Figures~\ref{fig:attention_cls} and~\ref{fig:attention_center} both visualize an image from the \texttt{Tench} class, the first class in ImageNet-1k, which represents a type of freshwater fish.
Figure~\ref{fig:attention_cls} shows the CLS-to-patch interaction maps.
Compared with ViT, VisionScreen produces more selective interaction patterns, reflecting the explicit rejection of low-relevance patches by Screening.
This suggests that VisionScreen can focus on visually relevant regions without distributing weights across all patches through softmax normalization.
Figure~\ref{fig:attention_center} shows the interaction maps from the center patch to all patch tokens.
In this visualization, we examine how the selected center token attends to other regions during token aggregation. ViT tends to distribute attention over a wide spatial range, sometimes assigning strong weights to distant regions such as image corners or to objects unrelated to the target class.
In the visualized example, ViT assigns relatively strong weights to fishing equipment, such as a fishing rod or reel, rather than to the fish itself.
In contrast, VisionScreen focuses more consistently on the fish region, indicating that it aggregates information from semantically relevant patches.
This behavior suggests that VisionScreen is less prone to relying on spurious correlations and instead captures more class-relevant visual evidence.

These visualizations qualitatively support the quantitative results, showing that the performance improvement of VisionScreen is accompanied by more selective and semantically meaningful patch aggregation behavior.

\subsection{Analysis of Spatial Softmask}

\begin{figure}
\centering
\includegraphics[width=0.95\linewidth]{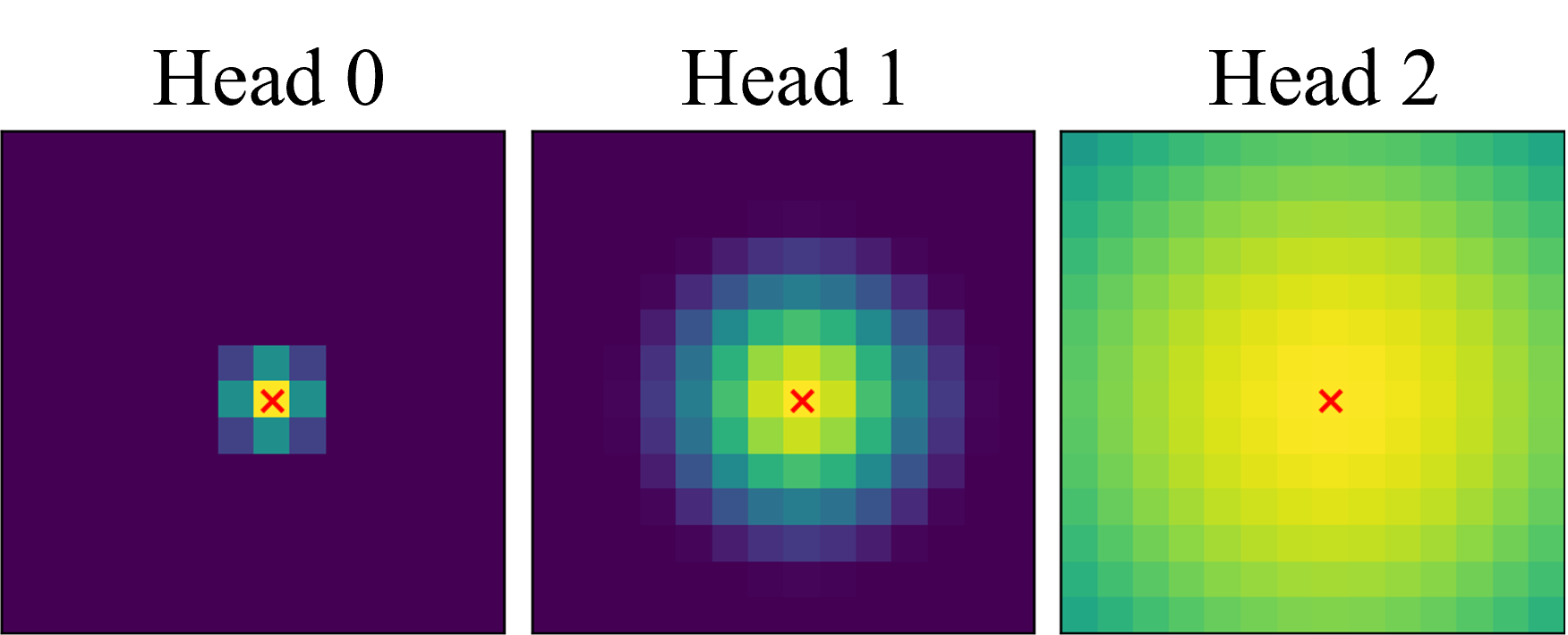}
\caption{
Visualization of Spatial Softmask in the final layer of VisionScreen.
Each panel shows the effective spatial range for a selected patch marked with a red $\times$.
Different heads exhibit different window sizes, ranging from local to global.
}
\label{fig:softmask}
\end{figure}

To further analyze the behavior of the proposed spatial softmask, we first visualize the spatial softmasks generated from the learned screening window parameters in the final layer of VisionScreen trained on ImageNet-1k.
Figure~\ref{fig:softmask} shows the spatial distribution of the softmask values for each head, with the selected patch marked by a red $\times$.
The screening window defined in Eq.~\ref{eq:screening_window} controls the spatial extent of the softmask in Eq.~\ref{eq:2d_softmask}.

As shown in Fig.~\ref{fig:softmask}, the spatial extents of the softmasks vary substantially across heads.
Some heads assign nonzero weights only to a relatively local region around the selected patch, while others cover a much broader area and capture more global context.
This result suggests that the spatial softmask does not impose a uniform spatial range across heads, but instead allows different heads to specialize in different spatial scales.
Smaller screening windows can capture fine-grained neighborhood information, whereas larger windows can incorporate long-range dependencies across the image.
Therefore, the learned screening windows allow VisionScreen to adaptively balance local and global information through its multi-head design.

\begin{figure}
\centering
\includegraphics[width=\linewidth]{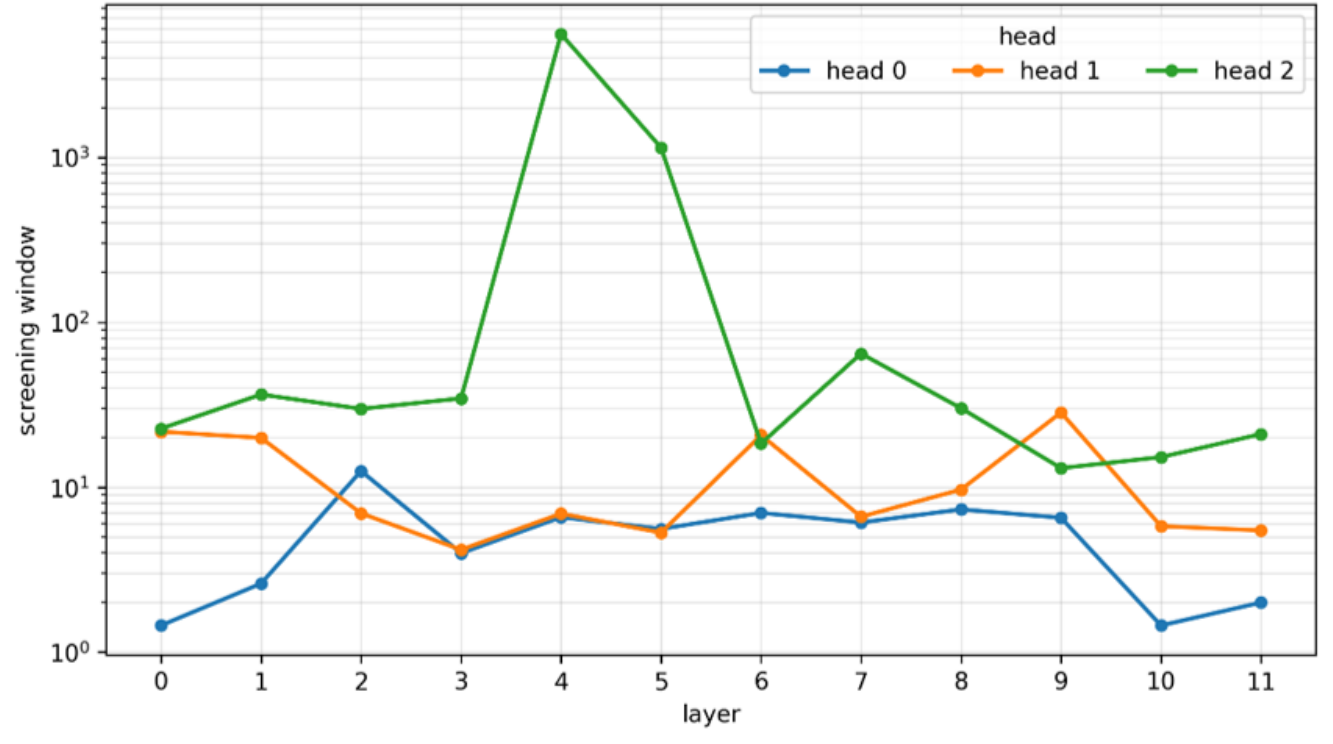}
\caption{
Learned screening window values across layers and heads.
The horizontal axis denotes the layer index, and the vertical axis denotes the screening window size.
Only a small number of heads have very large windows, while many heads preserve local or mid-range spatial aggregation even in deeper layers.
}
\label{fig:window_plot}
\end{figure}

We further analyze the learned screening window values across all layers and heads.
Fig.~\ref{fig:window_plot} shows the learned window size of each head, where the horizontal axis denotes the layer index and the vertical axis denotes the screening window size.
The windows in the final layer correspond to the visualization shown in Fig.~\ref{fig:softmask}.
Considering Figs.~\ref{fig:softmask} and~\ref{fig:window_plot} together with the definition of the spatial softmask in Eq.~\ref{eq:2d_softmask}, we observe that when the screening window becomes sufficiently large, approximately above 1000 in our setting, the softmask assigns nearly uniform weights across almost the entire patch grid.
In this case, the head behaves closer to global aggregation, similar to the behavior of standard ViT attention without an explicit spatial restriction.
However, Fig.~\ref{fig:window_plot} shows that only a small number of heads have such large window values.
Most heads retain spatially limited windows, indicating that VisionScreen does not simply degenerate into global aggregation.

Another observation is that both shallow and deep layers contain a mixture of local and global heads.
While deeper layers in ViT are often considered to mainly aggregate global information, our results suggest that local patch aggregation can remain important even in deeper layers.
This indicates that VisionScreen adaptively maintains multiple spatial scales throughout the network, rather than monotonically increasing the receptive range with depth.

\subsection{Ablation Study}


We further conduct an ablation study to analyze the effect of the gating mechanism in VisionScreen.
The gate consists of a linear projection followed by SiLU and Tanh activations, and its output is multiplied with the screened representation before the final projection.
This design follows the original Screening mechanism rather than being a newly introduced component in our method.
Nevertheless, since the gate directly modulates the aggregated features, we examine its effect on visual recognition performance.

Table~\ref{tab:gate_ablation} shows the ImageNet-1k top-1 accuracy with and without the gating mechanism.
Removing the gate decreases the accuracy from 72.5\% to 71.8\%, resulting in a 0.7 percentage point drop.
This result indicates that the gating mechanism contributes to the performance of VisionScreen.
We consider that the gate adaptively modulates the aggregated patch features on a channel-wise basis, suppressing less useful responses while emphasizing informative ones.
Since the model still outperforms ViT without the gate, we regard it as an auxiliary mechanism that enhances representational capacity rather than an essential component of Screening.

\begin{table}
\centering
\caption{
Ablation study on the gating mechanism in VisionScreen.
We report top-1 accuracy on ImageNet-1k.
}
\label{tab:gate_ablation}
\begin{tabular}{lc}
\toprule
Setting & ImageNet-1k Top-1 (\%) \\
\midrule
w/o Gate & 71.8 \\
w/  Gate & \textbf{72.5} \\
\bottomrule
\end{tabular}
\end{table}
\section{Conclusion}
\label{sec:conclusion}

We proposed VisionScreen, a vision model that extends the Screening mechanism to two-dimensional image patches.
By assigning an absolute relevance value to each patch pair, VisionScreen can explicitly reject low-relevance patches, unlike softmax attention, which aggregates features through relative weighting.
Experiments on ImageNet-1k and CIFAR-100 show that VisionScreen outperforms ViT-Tiny/16 with fewer parameters, while achieving more stable training and more selective patch interactions. 

Future work includes applying VisionScreen to dense prediction tasks and further analyzing its relevance patterns and patch selection behavior.

{
    \small
    \bibliographystyle{ieeenat_fullname}
    \bibliography{main}
}


\end{document}